\documentclass[a4paper,conference]{IEEEtran}
\usepackage[small]{caption}
\usepackage{amsmath}
\usepackage{booktabs}
\usepackage{algorithm}

\usepackage{graphicx}
\usepackage{subfigure}
\usepackage{float}
\usepackage{amsthm,amsmath,amssymb}
\usepackage{mathrsfs}
\usepackage{algpseudocode}
\usepackage{booktabs}
\usepackage{threeparttable}
\usepackage{multirow}
\usepackage{amsthm}
\usepackage{indentfirst}
\usepackage{colortbl}
\usepackage{tabularx}
\usepackage{arydshln}
\usepackage{xspace}
\usepackage{booktabs}
\usepackage{algorithm,algpseudocode}
\usepackage{epsfig}
\usepackage{graphicx}
\usepackage{amsmath}
\usepackage{amssymb}
\usepackage[pagebackref=true,breaklinks=true,letterpaper=true,colorlinks,bookmarks=false]{hyperref}
\usepackage{pbox}
\usepackage{capt-of}

\usepackage[normalem]{ulem}
\usepackage{multirow}

\usepackage{colortbl}

\usepackage{mathtools}
\usepackage{array}
\usepackage{CJK}
\usepackage{type1cm}
\usepackage{times}
\usepackage{footnote}
\usepackage[marginal]{footmisc}
\renewcommand{\thefootnote}{}

\ifCLASSINFOpdf
\else
\fi
\begin{document}

%
\title{A Neural Architecture Search based Framework for Liquid State Machine Design}

\author{\IEEEauthorblockN{Shuo Tian\IEEEauthorrefmark{1}, Lianhua Qu\IEEEauthorrefmark{1}, Kai Hu, Nan Li, Lei Wang and Weixia Xu\\
 College of Computer Science and Technology, National University of Defense Technology, Changsha, China\\
Email: tianshuo14@nudt.edu.cn, qulianhua14@nudt.edu.cn, arrowya@gmail.com}}

\newcommand\blfootnote[1]{%
  \begingroup
  \renewcommand\thefootnote{}\footnote{#1}%
  \addtocounter{footnote}{-1}%
  \endgroup
}

\maketitle

\begin{abstract}
Liquid State Machine (LSM), also known as the recurrent version of Spiking Neural Networks (SNN), has attracted great research interests thanks to its high computational power, biological plausibility from the brain, simple structure and low training complexity.
By exploring the design space in network architectures and parameters, recent works have demonstrated great potential for improving the accuracy of LSM model with low complexity.
However, these works are based on manually-defined network architectures or predefined parameters.
Considering the diversity and uniqueness of brain structure, the design of LSM model should be explored in the largest search space possible.
 
In this paper, we propose a Neural Architecture Search (NAS) based framework to explore both architecture and parameter design space for automatic dataset-oriented LSM model.
To handle the exponentially-increased design space, we adopt a three-step search for LSM, including multi-liquid architecture search with parallel and serial hierarchical topology, variation on the quantity of neurons in each liquid and parameters search such as percentage connectivity and excitatory neuron ratio within each liquid.  
Besides, we propose to use Simulated Annealing (SA) algorithm to implement the three-step heuristic search.
Three datasets, including image dataset of MNIST and NMNIST and speech dataset of FSDD, are used to test the effectiveness of our proposed framework.
Simulation results show that our proposed framework can produce the dataset-oriented optimal LSM models with high accuracy and low complexity.
The best classification accuracy on the three datasets is 93.2\%, 92.5\% and 84\% respectively with only 1000 spiking neurons, and the network connections can be averagely reduced by 61.4\% compared with a single LSM.
Moreover, we find that the total quantity of neurons in optimal LSM models on three datasets can be further reduced by 20\% with only about 0.5\% accuracy loss. 
\end{abstract}

\blfootnote{ \IEEEauthorrefmark{1}Shuo Tian and \IEEEauthorrefmark{1} Lianhua Qu are equal contribution.}

%
\IEEEpeerreviewmaketitle


\section{Introduction}

Spiking Neural Network (SNN) is recognized as the third generation artificial neural network aiming to achieve artificial intelligence by mimicking the computation and learning methods of brain \cite{schuman2017survey}. Compared with the general deep neural network, SNN has higher computing power \cite{maass1997networks}, and seems to be more reasonable in biology, since they can simulate the temporal information transmission between biological neurons.

Liquid State Machine (LSM), as a recurrent variant of SNN, is gaining popularity due to its intrinsic processing capability of spatio-temporal information, simple structure and low training complexity \cite{maass2002realtime,Maass2007liquid}. The LSM mainly consists of input layer with encoding neurons, a recurrent spike neural network called liquid and a set of readout neurons that extract liquid states from the liquid. Encoding neurons generally convert the input data into spike trains to the liquid.
Neurons in the liquid are randomly connected by percentage connectivity with fixed but stochastically weights.
Unlike the traditional machine learning neural network, only weights of connections from the liquid to readout neurons need training.
Despite its simple structure, LSM has shown its high computing power on various applications such as machine learning tasks, including speech recognition, image classification, and word recognition \cite{jalalvand2018on,verstraeten2005isolated,goodman2006spatiotemporal,grzyb2009facial}.  

Recently, many works are devoted to the performance enhancement of LSM models with low complexity \cite{ju2013effects,wijesinghe2019analysis,mi2019spatiotemporal}.
Parami \textit{et al.} analyze how the segmentation of a single large liquid to multiple smaller liquids affects the accuracy for LSM \cite{wijesinghe2019analysis}. These generated liquids are independent of each other, which can be evaluated in parallel. The experiment results show that the parallel ensemble method can effectively improve the accuracy without additional change for the simple structure of LSM. Moreover, for a given percentage connectivity, grouping a single LSM into several parallel small liquids can significantly reduce the available connections of LSM \cite{wijesinghe2019analysis}. This also reduces the storage requirement for LSM. 
Inspired by the form of connection in the biological nervous system, Mi \textit{et al.} propose to apply each small liquid in series to process spatio-temporal information \cite{mi2019spatiotemporal}.
However, to our best knowledge, no exploration has been carried out on hierarchical LSM models of multiple liquids with both parallel and serial connections.

The percentage connectivity which indicates the connection probability between the neurons within the liquid plays an important role in improving the accuracy of LSM \cite{wijesinghe2019analysis}. Too high or low connectivity will harm accuracy performance, which also suggests that there is an optimal percentage connectivity.  
Ju \textit{et al.} present a genetic algorithm-based approach to explore the parameters related to percentage connectivity in one liquid for higher classification accuracy \cite{ju2013effects}.
Reynolds \textit{et al.} \cite{reynolds2019intelligent} utilize an evolutionary algorithm to optimize the number of neurons and percentage connectivity on a single liquid for training accuracy. 
Zhou \textit{et al.} \cite{zhou2019evolutionary} use a covariance matrix adaptation evolution strategy to optimize three parameters, i.e., percentage connectivity, weight distribution and membrane time constant in one liquid. However, they all only perform parameter optimization in a single liquid and do not optimize the architectures of LSM.
Therefore great efforts are still needed to explore the potential of better architectures and parameters for LSM.

Neural Architecture Search (NAS) has achieved comparable performance to neural architectures by human experts in Recurrent Neural Network (RNN), Convolutional Neural Network (CNN) and even Graph Convolutional Network (GCN) \cite{peng2019learning, zoph2018Learning, zoph2016neural}.
It is natural to wonder if NAS could drive performance enhancement for LSM. 
In this paper, we introduce the NAS method to exploit the network architectures and parameters potential of LSM, and establish a framework for dataset-oriented LSM model design.
Specifically, we adopt a three-step search strategy, including multiple-liquid architecture search, neuron number variation, and parameter search for each liquid.
In step 1 of multiple-liquid architecture search, given the total number of neurons, the LSM model is divided averagely into several smaller liquids with parallel and serial hierarchical topology.
In step 2, neuron number variation explores the capability of each liquid with a different quantity of neurons.
For step 3, the parameter search of various percentage connectivity in each liquid and excitatory to inhibitory neuron ratio is applied to the optimal LSM model in step 2 for further optimization.
Moreover, for the complexity reduction, redundant neurons are proportionally reduced to minimize neuron numbers of the optimal model.

The main contribution of this work is as follows:
\begin{itemize}
\item We propose a NAS-based framework for designing dataset-oriented LSM model;
\item We define a complete whole search space for LSM and use a three-step search to explore multiple liquids architecture, neuron number and parameters of each small liquid;
\item We propose to use Simulated Annealing based algorithm to handle the three-step search in the framework;
\item We analyze the trade-off between total quantity of neurons and accuracy to reduce the complexity of hardware implementation.
\end{itemize}

The best classification accuracy on MNIST, NMNIST and FSDD of the optimal models is 93.2\%, 92.5\% and 84\% with only 1000 spiking neurons, and the network connections can be reduced by 61.4\% averagely compared with a single liquid. Additionally, the total neuron number of optimal LSM models can be further reduced by 20\% with only about 0.5\% accuracy loss. 

\section{Related work}
The performance of LSM-based learning model is mainly determined by the network structure and hyperparameters which are related to the network scale and connectivity patterns in supervised learning.
Inspired by the mammalian neocortex, Maass \textit{et al.} first propose the Liquid State Machine (LSM) as a feature filter using randomly interlinked spiking neurons.
Many experiments have demonstrated that LSM can be used to project complex input patterns to high-dimension linearly-separable liquid states.
The projected liquid states can be fed into the traditional machine learning method to achieve the classification tasks.
Though the accuracy of LSM model could be improved by increasing the number of neurons, the network complexity and computation cost are also raised.
Therefore, recent works are mainly focused on improving accuracy with low complexity.

Ju \textit{et al.} present a genetic algorithm-based approach to evolve the liquid filter from a minimum structure without connections to an optimized kernel with a minimal number of synapses and high classification accuracy \cite{ju2013effects}.
The weight distribution of the resulted LSM model is similar in shape with cortical circuitry, which indicates the potential for exploring the percentage connectivity of LSM.
Though Ju \textit{et al.} observed that there were many potential connection topologies, their explorations were limited by the computing resources at that time.

Parami \textit{et al.} propose an ensemble method to improve classification accuracy and reduce the connections of LSM \cite{wijesinghe2019analysis}.
In Parami's method, a single large LSM is divided into several small liquids with the same number of neurons.
The results show that a different number of liquids may lead to different accuracy on the same dataset which indicates the existence of an optimum number of liquids for different applications \cite{wijesinghe2019analysis}. 
However, Parami \textit{et al.} only use the method of equal division.
The potential of different neuron number and percentage connectivity in each small liquid for LSM is not exploited.

Mi \textit{et al.} propose a canonic neural network model for spatiotemporal pattern recognition by combining a liquid subnetwork and a decision-making subnetwork \cite{mi2019spatiotemporal}.
The liquid subnetwork in Mi's proposal is in a hierarchical structure with a number of serial liquids.
The analysis simulations show that the dominating components of neuronal responses progress from high to low frequencies along with the hierarchy layer.
This implies that a hierarchical liquid can be used to discriminate input patterns with different frequencies.
However, no realistic applications are performed based on the proposed hierarchical liquid model, rather only a single liquid is evaluated.
In conclusion, models with hierarchical multi-layer are still poorly understood, and the optimal network parameters such as percentage connectivity and topology have not been explored for LSM.

\begin{figure*}[!t]
\centering
\includegraphics[width=1.0\linewidth]{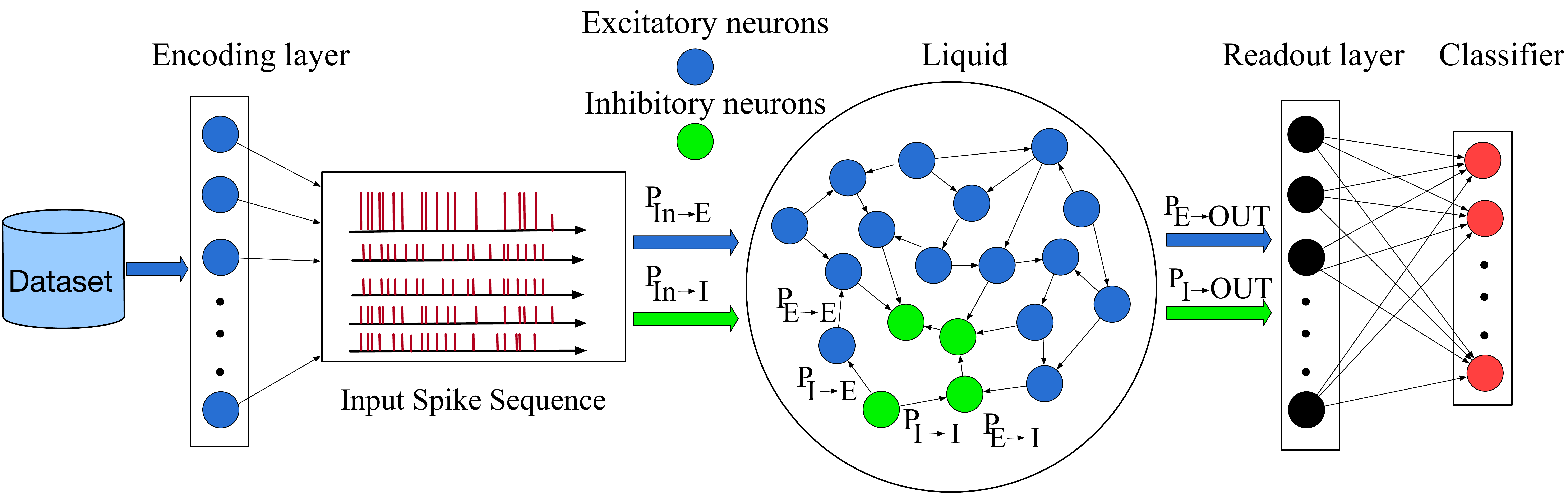}
\caption{The schematic of the classical LSM learning model.}
\label{LSM_classical}
\end{figure*}

Since manually adjusting network parameters is a tedious and heavy task, increasing efforts have been paid to optimize parameters of LSM by automatic search methods.
Reynolds \textit{et al.} \cite{reynolds2019intelligent} utilize an evolutionary algorithm to optimize the number of neurons and percentage connectivity on one liquid for training accuracy.
Zhou \textit{et al.} \cite{zhou2019evolutionary} propose a covariance matrix adaptation evolution strategy to optimize three parameters, i.e., percentage connectivity, weight distribution and membrane time constant for a single liquid on three datasets. 
However, both works of Reynoulds \textit{et al.} \cite{reynolds2019intelligent} and Zhou \textit{et al.} \cite{zhou2019evolutionary} only search the parameters of a single liquid.
The automatical optimization of test accuracy in the architecture of LSM such as multiple liquids and their parameters remains to be studied.

The purpose of NAS design is to obtain the advanced neural network architectures with less or without human efforts under the legitimate calculation budgets.
Neural architecture search (NAS) is first proposed by Zoph \textit{et al.} \cite{zoph2016neural}, which automates the process of discovering promising neural architectures.
Reinforcement learning (RL) \cite{zoph2016neural} and evolutionary algorithms \cite{real2019regularized} are currently two mainstream methods to NAS. The original NAS \cite{zoph2016neural} learns an RNN controller to sample a better architecture. Real \textit{et al.} \cite{real2019regularized} introduce evolutionary algorithm as a controller to obtain promising architectures for CNN. Wang \textit{et al.} \cite{wang2019particle} use particle swarm optimization (PSO) algorithm to search for advanced CNN architectures. 
In areas such as image classification tasks \cite{zoph2018Learning} and semantic image segmentation \cite{liu2019auto}, the architecture of automatic design has been superior to the well-known manual architecture. There have also been some attempts in NAS for motion recognition from RGB data \cite{peng2019video} and non-Euclidean data \cite{peng2019learning}. However, little work has provided the idea of NAS into the LSM architecture search process. To our best knowledge, this paper is the first work to use NAS to search neural architecture as well as network parameters for LSM.

\section{Methods}
 
\subsection{LSM model}

A typical LSM-based learning model mainly consists of three components: encoding layer, liquid and readout layer.
As shown in Figure \ref{LSM_classical}, the input data is firstly converted into the spike sequence by encoding layer following specific encoding schemes, such as rate coding, time coding and phase coding.
Then the encoding neurons that only contain excitatory neurons are randomly and sparsely connected to the neurons in the liquid.
The liquid of randomly inter-linked spiking neurons is used to project the low-dimension input information to a high-dimension linearly-separable liquid state \cite{Maass2007liquid}.

Both excitatory and inhibitory neurons in the liquid are simulated using the leaky integrate-and-fire model \cite{diehl2015unsupervised}. 
The liquid contains the connections from excitatory to excitatory neurons ($E \rightarrow E$), excitatory to inhibitory neurons ($E \rightarrow I$), inhibitory to excitatory neurons ($I \rightarrow E$) and inhibitory to inhibitory neurons ($I \rightarrow I$).
The percentage connectivity ($P_{E \rightarrow E}$, $P_{E \rightarrow I}$, $P_{I \rightarrow E}$, $P_{I \rightarrow I}$) within the liquid are usually the same with those used in Diehl \textit{et al.} \cite{diehl2015unsupervised}. The percentage connectivity of input to liquid ($P_{In \rightarrow E}$, $P_{In \rightarrow I}$) are set to 0.1, 0, respectively. The ratio of excitatory to inhibitory neurons stays the same with Wehr \textit{et al.}\cite{wehr2003balanced}. Since the ratio of excitatory and percentage connectivity ($P_{E \rightarrow E}$, $P_{E \rightarrow I}$, $P_{I \rightarrow E}$, $P_{I \rightarrow I}$) within the liquid have an important influence for classification ability of LSM, we mainly optimize the five parameters for the liquid in this paper.
The synapse for transmitting spikes can be modeled by an exponentially-decaying current source as in Diehl \textit{et al.} \cite{diehl2015unsupervised}.

By simulating the liquid with the input spike sequence, we can get the liquid state corresponding to the presented input example.
In this work, we use the normalized spike number of every excitatory neuron in the liquid as the liquid state after presenting the input spike sequence.
Therefore, the readout layer is responsible for recording the spike number of sampled neurons in the liquid and process the spike number into feature vectors.
The classifier is a simple perceptron with the number of output neurons equal to the classes needed to be discriminated against.
Stochastic gradient descent is used for the training of the perceptron.
\begin{figure*}[!t]
\centering
\includegraphics[width=1\linewidth]{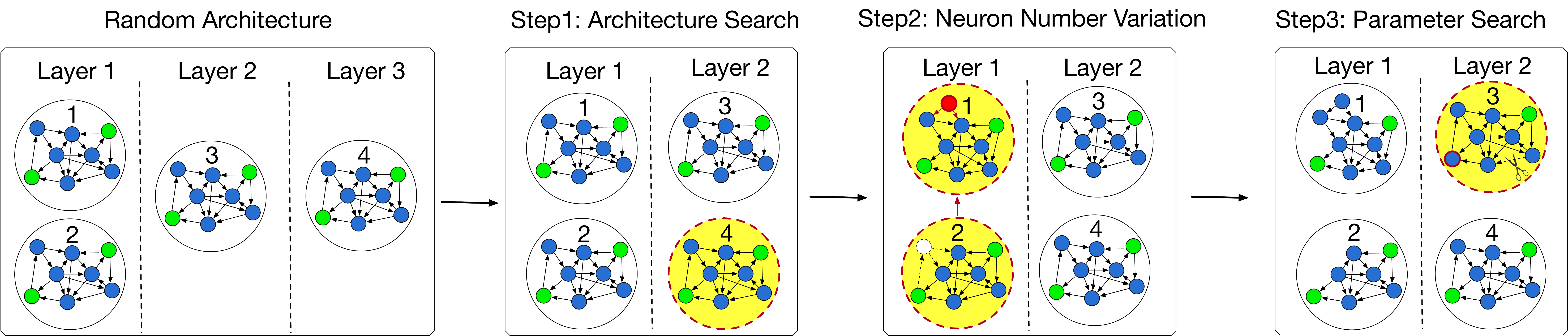}
\caption{The three-step search for LSM. The first step is to explore optimal LSM architecture. In the second step, we optimize the number of neurons in each liquid based on the optimal model in step 1. Finally, we optimize the internal parameters of each liquid on account of the optimal model in step 2.}
\label{LSM_stage}
\end{figure*}

\subsection{NAS-based framework}

\begin{figure}[H]
\centering
\includegraphics[width=1.0\linewidth]{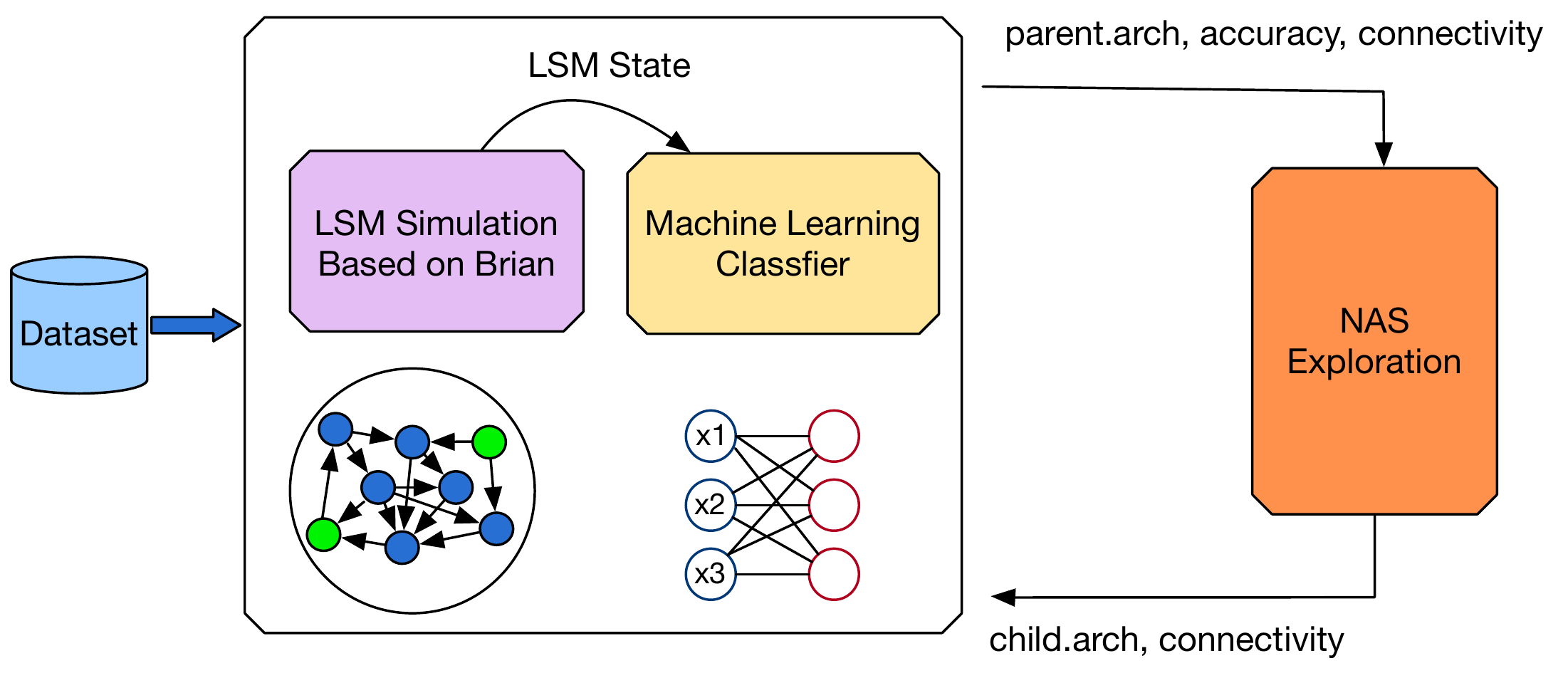}
\caption{The NAS-based framework for LSM design.}
\label{search_framework}
\end{figure}
The framework of NAS for LSM is shown in Figure \ref{search_framework}. The dataset is converted into the spike version and fed into the \textit{Brian} simulator \cite{goodman2009the,briansimulator} with the searched structure. After simulation, the liquid state is extracted from the simulator output and fed into the classifier for training and test. The test accuracy including the existing $parent$ architecture ($parent.arch$), $connectivity$, etc will be given to the NAS exploration module. Then NAS explores and returns a $child$ architecture as well as the percentage connectivity, etc to the \textit{Brian} simulator. This process is repeated many times until the search is completed or the optimal model is found.

\subsection{Three-step neural network search for LSM}

We use a three-step search for the optimal LSM. 
As shown in Figure \ref{LSM_stage}, the first step is to explore optimal LSM architecture.
In the second step, based on the optimal architecture, we optimize the number of neurons in each liquid based on the optimal model in step 1. Finally, we have optimized the internal parameters of each liquid on account of the optimal model in step 2.

A Simulated Annealing (SA) algorithm \cite{van1987Simulated} is used in every step as demonstrated in Algorithm \ref{alg:A}.
The SA is one kind of greedy algorithm \cite{edmonds1971matroids}, but a random factor is introduced in the search process. The SA algorithm starts from a higher initial temperature, and with the continuous decrease of temperature, it gradually finds the optimal solution. When updating the current feasible solution, it will accept a solution that is worse than the current one with a certain probability, which is possible to make the current solution jump out of the local optimum and reach the global optimum.

\begin{algorithm}
\caption{Simulated Annealing based three-step search algorithm}
\label{alg:A}
\begin{algorithmic}
\State$T_{initial}$ //{initiate temperature}
\State$Tmin$ //{minimum value of temperature}
\State$T$ //{current temperature}
\State$k$ //{times of internal cycle} 
\State$t$ //{times of $T_{initial}$ decline} 
\State $history\gets\varnothing$ //{Will contain all the models.}
\State $parent.arch\gets$RandomArchitecture()
\State $parent.acc\gets$Train and Evaluate on test set
\State add $parent$ to history
\State $T = T_{initial}$
\While{$T > Tmin$}
\For i in range(k):
\State $chlid.arch\gets$Disturb ($parent.arch$)
\State $chlid.acc\gets$Train and Evaluate on test dataset
\State add $child$ to history
\If{$child.acc > parent.acc$}
\State parent = child
\Else  
\State $p = exp(-(parent.acc - child.acc)/T)$
\State$ r = random(0,1)$
\If{$ r < p$}
\State parent = child
\EndIf
\EndIf
\EndFor
\State $t + = 1$
\State $T = T_{initial} / (t +1) $
\EndWhile\label{euclidendwhile}
\State \textbf{return} highest-accuracy model in $history$
\end{algorithmic}
\end{algorithm}

\paragraph{Architecture search for LSM}
\begin{figure}[!t]
\centering
\includegraphics[width=0.9\linewidth]{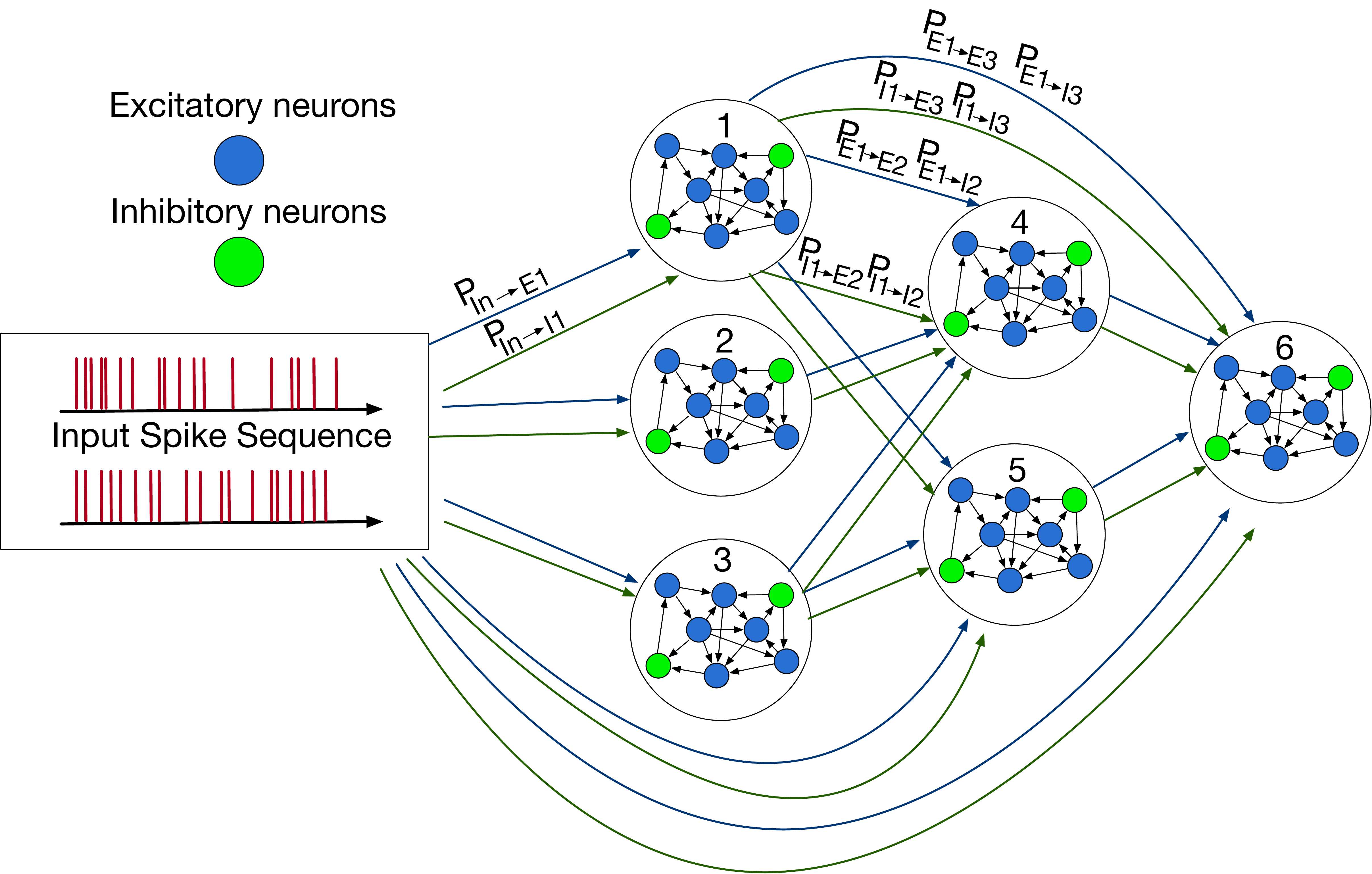}
\caption{The search space designed for LSM. There are six individual liquids in a three-layer LSM for the figure above. The search space consists of the forward connected layers between parallel and cascade liquids. All liquids including the input in the previous layers are connected to the liquids in the post layer.}
\label{LSM_multilayer}
\end{figure}
Specifically, our first step is searching the optimal architecture. 
Parami \textit{et al.} divide a large liquid into multiple smaller units \cite{wijesinghe2019analysis}. These small liquids work in parallel separately within one layer. Mi \textit{et al.} form each small liquid in series to process the spatio-temporal information \cite{mi2019spatiotemporal}. In this paper, our architecture search space mainly combine the above parallel and series architectures, which is also similar to feedforward hierarchical architectures in deep learning \cite{lecun2015deep}.
As shown in Figure \ref{LSM_multilayer}, the architecture search space is
represented by the forward connected layers between parallel and cascade liquids. 

There are six individual liquids in a three-layer LSM in Figure \ref{LSM_multilayer}. Multiple liquids in one layer work in parallel separately and are not connected with each other.
Liquids in the pre-layer including the input layer have the opportunity to connect to all liquids in the post layer, but not the other way around. 
The percentage connectivity from $i$ layer to $j$ layer ($P_{Ei \rightarrow Ej}$, $P_{Ei \rightarrow Ij} $, $P_{Ii \rightarrow Ej} $, $P_{Ii \rightarrow Ij}$) is 0.1, 0, 0.1, 0, respectively.
The total number of neurons for a large liquid ($N_{total}$) is divided into $N_{group}$ small units and the quantity of each small liquid is $N_{total} / N_{group}$. The percentage connectivity and excitatory neurons ratio of each liquid are the same as the single LSM discussed in section 3, subsection A.

First we randomly construct an architecture recorded as the $parent$ and get its accuracy as shown in Algorithm \ref{alg:A}. Then we save the parent model in history list. After that the $child$ model is generated by randomly disturbing from the parent model. As shown in the step 1 Figure \ref{LSM_stage}, a certain liquid in one layer (liquid 4 in Layer 3) is randomly selected, and the liquid is stochastically added to another layer (Layer 2) of LSM.

Then we retrain the child model of final readout layer to get its accuracy and add the child model to history list. If the performance of child is better than parent model, then parent model is replaced by the child as demonstrated in Algorithm \ref{alg:A}. Otherwise, the child will replace parent model with a dynamic probability $p$.
When the current temperature $T$ falls into the lowest $Tmin$, the optimal architecture achieving the highest accuracy found by SA can be returned from the history list.
\paragraph{Neuron number variation}
In the second step, we also use SA algorithm to explore the optimal number of neurons in each small liquid. The parent model initialized here is not the random architecture, but the optimal architecture returned from history list in first step. The random disturbance added here are shown in the step 2 Figure \ref{LSM_stage}. Again, we randomly select a liquid (liquid 2) and the amount of disturbance range $M$ ($0 < M < N_{total} / N_{group}$).
Subsequently, the number of neurons inside the liquid (liquid 2) is randomly removed by number $m$ ($0 < m < M$), and another liquid (liquid 1) is stochastically chosen to add the reduced $m$ neurons.
Then the overall number of neurons remains unchanged.
The rest operations are the same as the first step as shown in Algorithm \ref{alg:A}, and finally the history list returns the best performance number of neurons in each liquid for the optimized architecture in step 1.
\begin{table}[!t]
\renewcommand{\arraystretch}{1.6}
\caption{Parameters range for each liquid}
\label{parameter}
\centering
\begin{tabular}{p{0.5\linewidth}c}
\hline
parameter & range \\
\hline
 Excitatory neurons ratio & $[0.0, 0.9]$ \\
$ P_{E \rightarrow E}$  & $[0.0, 0.9]$ \\
$P_{E \rightarrow I}$  & $[0.0, 0.9]$ \\
$P_{I \rightarrow E}$  & $[0.0, 0.9]$ \\
$P_{I \rightarrow I}$  & $[0.0, 0.9]$ \\
\hline
\end{tabular}
\end{table}
\paragraph{Parameters search for LSM}

Similar to the second step, we use SA algorithm to optimize the five internal parameters (as shown in Table \ref{parameter}) of each liquid. We limit the range of each parameter as shown in Table \ref{parameter}. When the parameter varies to a minus, the value is 0. And when it changes to 1, we set it to 0.9. The parent model initialized here is the optimal model returned in step 2. The random disturbance here is shown in step 3 Figure \ref{LSM_stage}. It mainly randomly selects a liquid (liquid 3), and stochastically chooses one of five internal parameters, i.e., excitatory neurons ratio and percentage connectivity ($P_{E \rightarrow E}$, $P_{E \rightarrow I}$, $P_{I \rightarrow E}$, $P_{I \rightarrow I}$), and adds the variation $\delta$ ($\delta$ is 0.1 or - 0.1). As shown in step 3 Figure \ref{LSM_stage}, the ratio of excitatory neurons is increased. The clipped dashed line indicates that the percentage connectivity $P_{E \rightarrow E}$ is reduced. Eventually, the highest performace optimal model of three-step search is returned from the history list.


\section{Experiments and results}
\subsection{Experimental setup and dataset}
The LSM model is simulated on the open-source SNN simulator of \textit{Brian} \cite{goodman2009the,briansimulator}. Three datasets of different properties are selected for the evaluation of our proposed framework, including MNIST \cite{mnist}, NMNIST \cite{orchard2015converting}, and FSDD \cite{fsdd}.
MNIST dataset contains 60000 training and 10000 testing grayscale handwritten number images belonging to ten categories of `0-9'.
NMNIST is the Dynamic Vision Sensor (DVS) of MNIST dataset, containing the same number of examples presented by events.
FSDD (Free Spoken Digit Dataset) is a free open speech dataset including recordings of spoken digits `0-9' in wav files at 8kHz.
It mainly consists of 2,000 audio recordings from 4 speakers (50 of each digit per speaker) in English pronunciations.
We separate the 2,000 recordings into a training dataset of 1,600 and a testing dataset of 400.
\subsection{Implementation details}
Except for NMNIST dataset, both MNIST and FSDD are converted to the Poisson spike sequence with the spike rate proportional to the intensity of input channel.
Specifically, the input of MNIST dataset is coded by rate coding scheme where the absolute values of pixels are converted to spike rate by input Poisson spike generator in \textit{Brian} \cite{mochizuki2014analog,briansimulator}.
The maximum spiking rate, corresponding to the highest grayscale value of 255, is 63.75 Hz.
For FSDD dataset, the Mel-scale Frequency Cepstral Coefficients (MFCC) feature is first extracted using the Python module of Lyons \textit{et al.} \cite{mfcc_python}. 
Then the extracted MFCC feature is coded by the same rate coding scheme with the Poisson distribution.
We use the first 10000 examples of a full training dataset of MNIST and NMNIST for training.
For FSDD dataset, the full training dataset of 1,600 examples is used for training. 
The test dataset is not seen by the training procedure, and accuracy is tested on the full test dataset on three datasets.

The spike history produced by the \textit{Brian} simulator is firstly processed into liquid state vectors.
The liquid state vectors and the corresponding labels are used for the training of classifier. 
Normal distribution is used for weight initialization and fixed during the search process.
The total neuron number in LSM model is fixed at 1000, the same as Parami \textit{et al.} \cite{wijesinghe2019analysis}. 

\subsection{Results}
In this subsection, we present the test accuracy of the searched optimal model after each search step, and the baseline accuracy of LSM model with average-divided parallel liquids \cite{wijesinghe2019analysis}.
\paragraph{MNIST}
\begin{figure}[!t]
\centering
\includegraphics[width=0.9\linewidth]{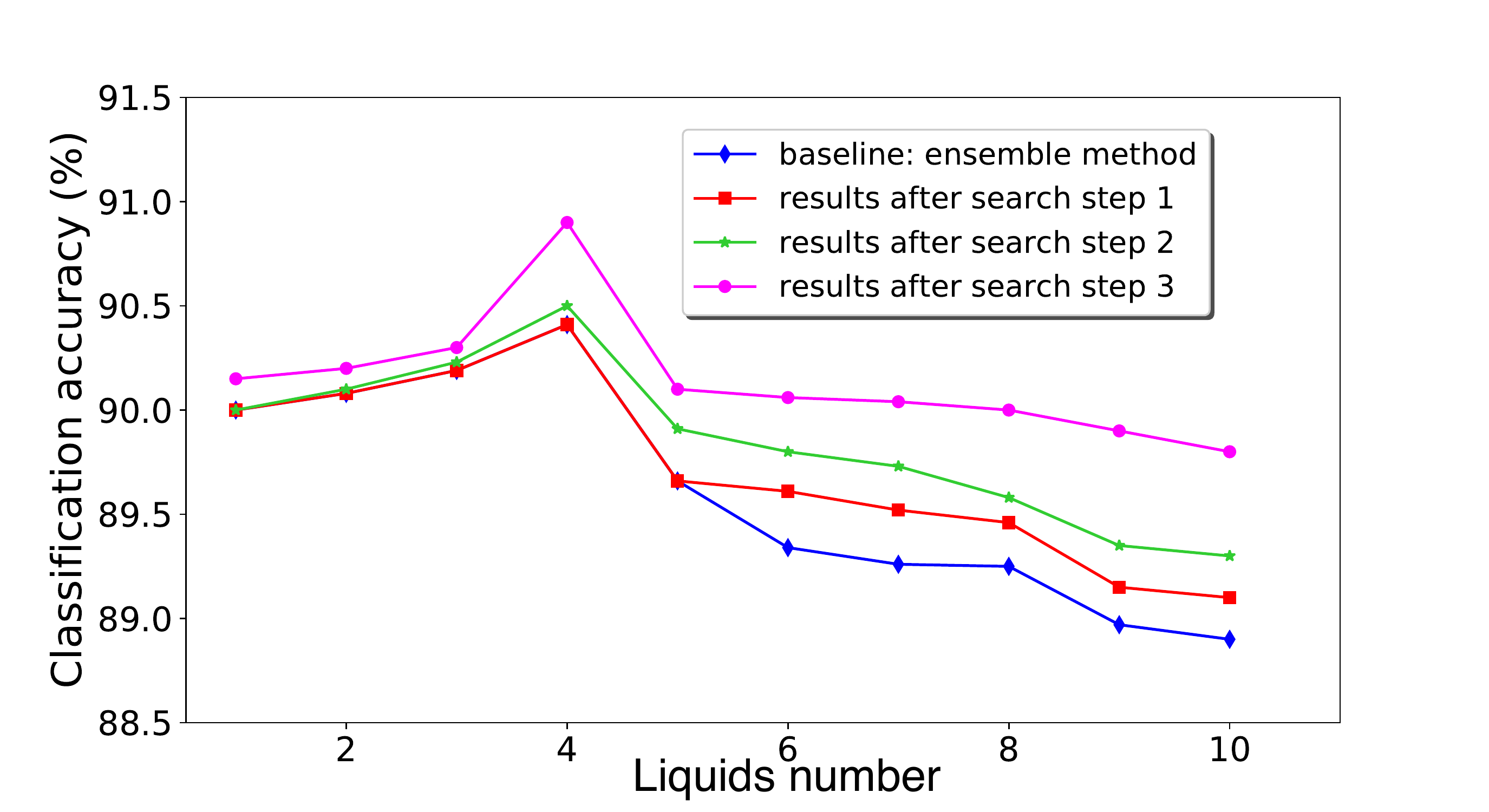}
\caption{The classification accuracy on MNIST dataset.}
\label{accuracy_mnist}
\end{figure}
Figure \ref{accuracy_mnist} shows the test accuracy on MNIST dataset for baseline \cite{wijesinghe2019analysis} and optimal model in each step under different number of liquids (from 1 to 10).
It can be seen that our proposed search framework could achieve accuracy improvement after every search step for different liqiuds number.
For the step of multiple-liquid architecture search (step 1), we find that parallel architecture may be the best for LSM model with less than five liquids.
While for more than five liquids, our hierarchical architecture with both parallel and cascade liquids can outperform the pure parallel one.
For example, the two-layer six liquids LSM model, i.e. four liquids in the first layer and two liquids in the second layer, achieves higher accuracy (89.6\%) than the single-layer model with six parallel liquids (89.3\%), which verifies the effectiveness of architecture search in step1.

This is mainly because for the number of liquids $N_{group}$, variable architectures in search space are $2^{N_{group} -1}$, when the number of liquids for LSM is small, there are fewer network architectures to choose from, and it is difficult to find a better architecture. As the number of $N_{group}$ increases, more network architectures can be selected, which is easier to find a better architecture than the pure parallel model. 

In step 2, for the first time, we find that the accuracy of optimal architectures could be improved by varying the neuron number for each individual liquid.
The reason may be that liquids with different neuron number are sensitive to different input patterns, which could enhance the state separation of the whole LSM model.  

For step 3, further accuracy improvement could be achieved by searching in the parameter space. This may because that different optimization parameter of each small liquid could affect the activation level for LSM, thereby improving the entire classification ability of LSM.
Therefore, it is necessary to optimize the parameters for a specific LSM model. The best LSM model for MNIST is the parallel four liquids as shown in Table \ref{architecture_mnist}.
\begin{table}[H]
\renewcommand{\arraystretch}{1.5}
\caption{The optimal LSM model for MNIST dataset}
\label{architecture_mnist}
\centering
\begin{tabular}{p{0.05\linewidth}p{0.1\linewidth}p{0.1\linewidth}ccccc}
\hline
Layer & Neuron number & Excitatory ratio & $E \rightarrow E$ & $E \rightarrow I$& $I \rightarrow E$ &$I \rightarrow I$\\
\hline
1 & 382 & 0.9 & 0.4 & 0.4 & 0.6 & 0.0 \\
1 & 99 & 0.9 & 0.4 & 0.4 & 0.5 & 0.0 \\
1 & 388 & 0.8 & 0.3 & 0.4 & 0.5 & 0.1 \\
1 & 131 & 0.8 & 0.4 & 0.4 & 0.5 & 0.0 \\
\hline
\end{tabular}
\end{table}

\paragraph{NMNIST and FSDD}

\begin{figure}[H]
\centering
\includegraphics[width=0.9\linewidth]{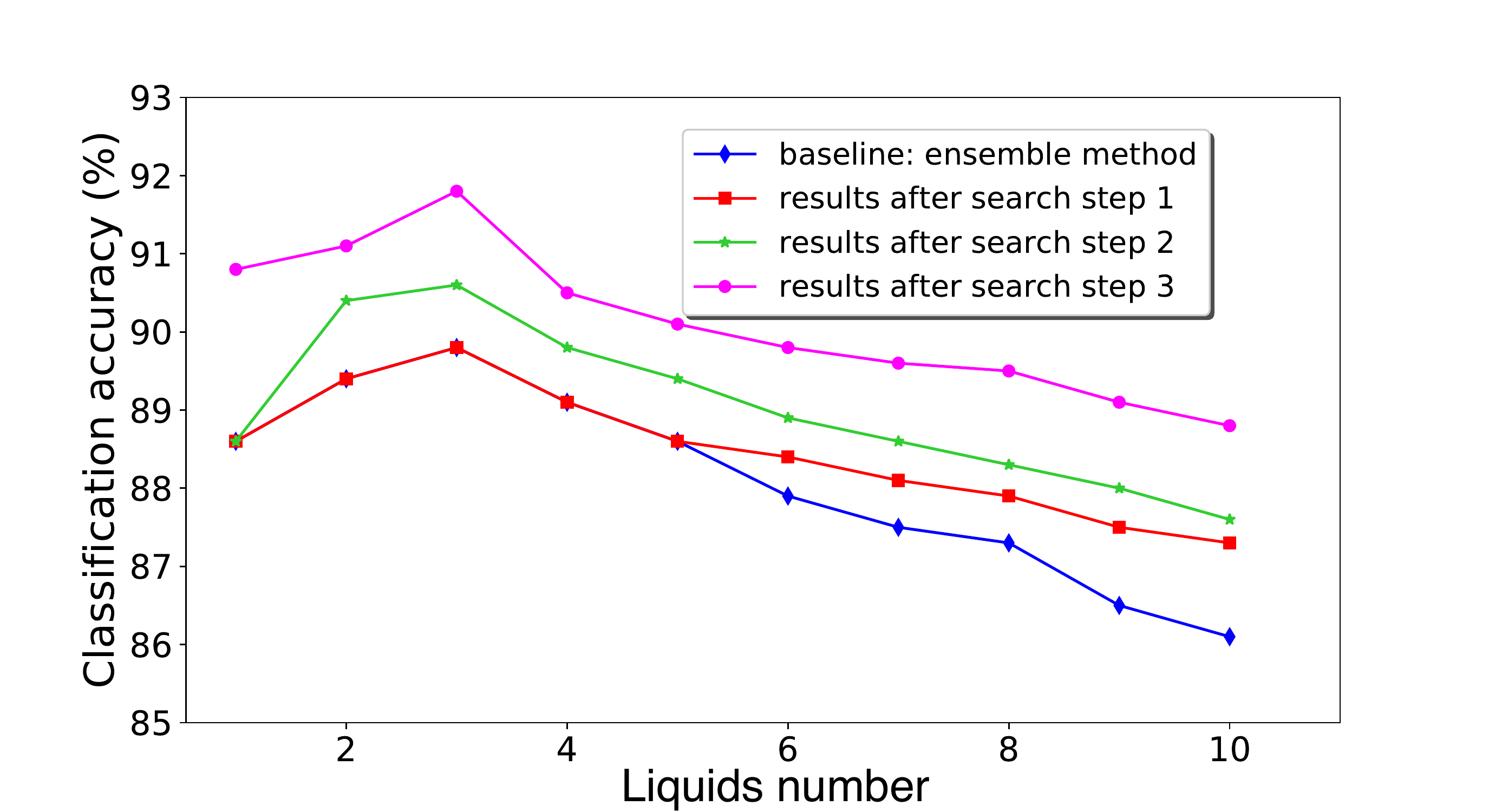}
\caption{The classification accuracy on NMNIST dataset.}
\label{accuracy_nmnist}
\end{figure}
\begin{figure}[H]
\centering
\includegraphics[width=0.9\linewidth]{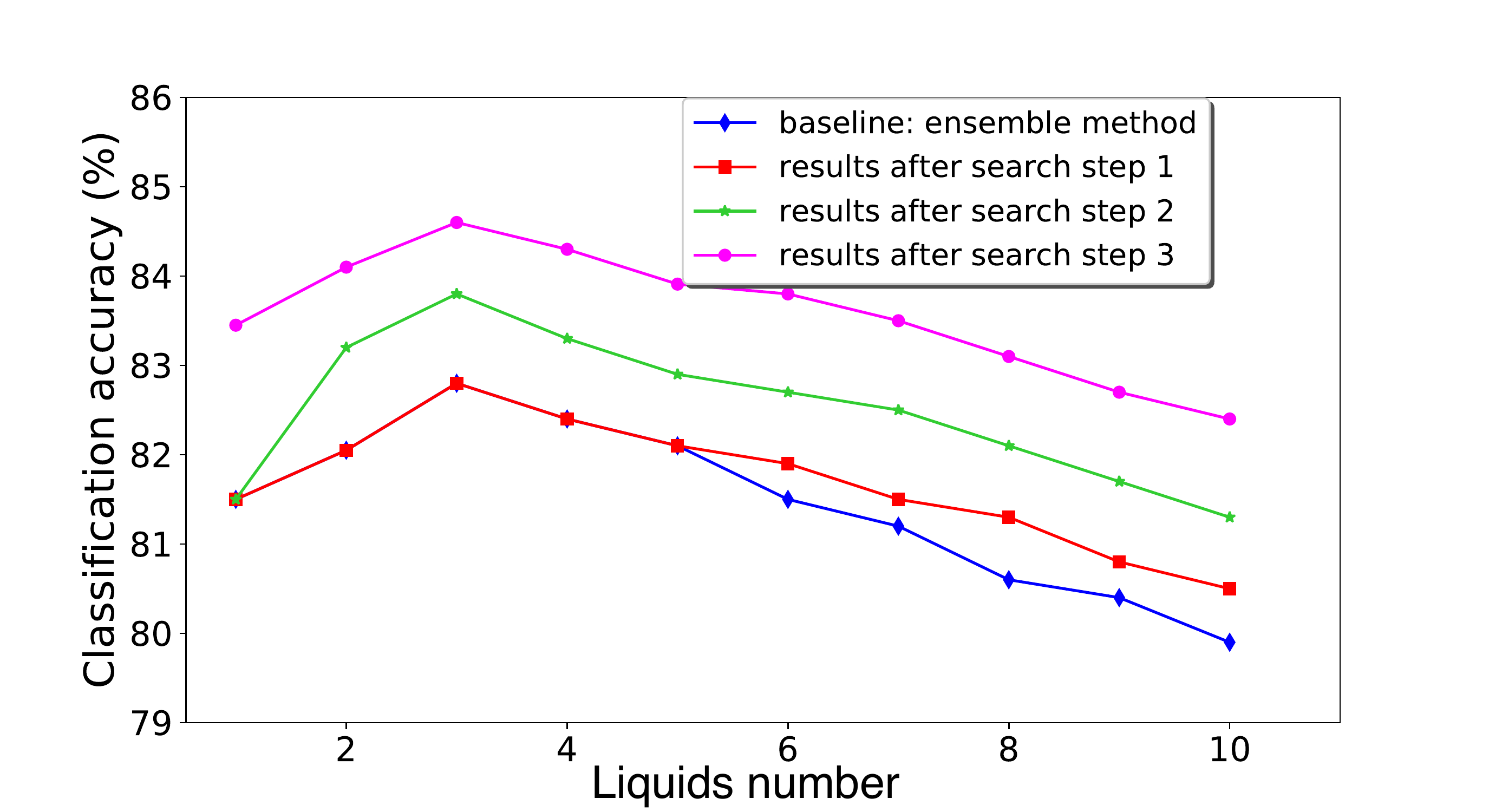}
\caption{The classification accuracy on FSDD dataset.}
\label{accuracy_fsdd}
\end{figure}
Figures \ref{accuracy_nmnist}, \ref{accuracy_fsdd} show the test accuracy on NMNIST and FSDD dataset.
It can be seen that the accuracy can be improved after every search step.
Similarly, when the number of liquids is less than five, pure parallel architecture is found to be the optimal one.
While, with more than five liquids, hierarchical architecture is found to be superior to pure parallel architecture \cite{wijesinghe2019analysis}. 
Different from MNIST dataset, the best network architecture lies in the three parallel liquids for NMNIST (as shown in Table \ref{architecture_nmnist1}) and FSDD (as shown in Table \ref{architecture_fsdd1}) dataset.
\begin{table}[H]
\renewcommand{\arraystretch}{1.5}
\caption{The optimal LSM model for NMNIST dataset}
\label{architecture_nmnist1}
\centering
\begin{tabular}{p{0.05\linewidth}p{0.1\linewidth}p{0.1\linewidth}ccccc}
\hline
Layer & Neuron number & Excitatory ratio & $E \rightarrow E$ & $E \rightarrow I$& $I \rightarrow E$ &$I \rightarrow I$\\
\hline
1 & 75 & 0.9 & 0.5 & 0.5 & 0.5 & 0.1 \\
1 & 505 & 0.8 & 0.4 & 0.3 & 0.4 & 0.0 \\
1 & 420 & 0.8 & 0.2 & 0.4 & 0.3 & 0.0 \\
\hline
\end{tabular}
\end{table}

\begin{table}[H]
\renewcommand{\arraystretch}{1.5}
\caption{The optimal LSM model for FSDD dataset}
\label{architecture_fsdd1}
\centering
\begin{tabular}{p{0.05\linewidth}p{0.1\linewidth}p{0.1\linewidth}ccccc}
\hline
Layer & Neuron number & Excitatory ratio & $E \rightarrow E$ & $E \rightarrow I$& $I \rightarrow E$ &$I \rightarrow I$\\
\hline
1 & 134 & 0.9 & 0.4 & 0.4 & 0.4 & 0.1 \\
1 & 382 & 0.9 & 0.5 & 0.6 & 0.6 & 0.1 \\
1 & 484 & 0.9 & 0.4 & 0.2 & 0.4 & 0.0 \\
\hline
\end{tabular}
\end{table}

\begin{table}[H]
\renewcommand{\arraystretch}{1.5}
\caption{The optimal LSM model with eight liquids for NMNIST dataset}
\label{architecture_nmnist}
\centering
\begin{tabular}{p{0.05\linewidth}p{0.10\linewidth}p{0.1\linewidth}cccc}
\hline
Layer & Neuron number & Excitatory ratio &  $E \rightarrow E$ &  $E \rightarrow I$ &  $I \rightarrow E$ &  $I \rightarrow I$ \\
\hline
1 & 62 & 0.9 & 0.5 & 0.5 & 0.6 & 0.0 \\
1 & 126 & 0.8 & 0.4 & 0.6 & 0.5 & 0.0 \\
1 & 70 & 0.9 & 0.3 & 0.5 & 0.5 & 0.1 \\
1 & 230 & 0.9 & 0.4 & 0.5 & 0.6 & 0.0 \\
1 & 98 & 0.8 & 0.4 & 0.4 & 0.4 & 0.1 \\
1 & 104 & 0.9 & 0.4 & 0.4 & 0.7 & 0.0\\
2 & 152 & 0.7 & 0.4 & 0.4 & 0.5 & 0.1 \\
2 & 158 & 0.9 & 0.3 & 0.4 & 0.4 & 0.0 \\
\hline
\end{tabular}
\end{table}

\begin{table}[H]
\renewcommand{\arraystretch}{1.5}
\caption{The optimal LSM model with eight liquids for FSDD dataset}
\label{architecture_fsdd}
\centering
\begin{tabular}{p{0.05\linewidth}p{0.1\linewidth}p{0.1\linewidth}cccc}
\hline
Layer & neuron number & Excitatory ratio &  $E \rightarrow E$&  $E \rightarrow I$ &  $I \rightarrow E$ &  $I \rightarrow I$ \\
\hline
1 & 134 & 0.9 & 0.4 & 0.4 & 0.6 & 0.0 \\
1 & 129 & 0.9 & 0.4 & 0.4 & 0.5 & 0.0 \\
1 & 148 & 0.9 & 0.4 & 0.4 & 0.5 & 0.0 \\
1 & 208 & 0.9 & 0.4 & 0.4 & 0.5 & 0.0 \\
1 & 42 & 0.9 & 0.4 & 0.4 & 0.5 & 0.0 \\
1 & 110 & 0.9 & 0.4 & 0.4 & 0.5 & 0.0\\
2 & 66 & 0.9 & 0.4 & 0.4 & 0.5 & 0.0 \\
2 & 163 & 0.9 & 0.3 & 0.4 & 0.5 & 0.0 \\
\hline
\end{tabular}
\end{table}

In particular, it is worth noticing that the accuracy of eight liquids with a two-layer hierarchical architecture after the optimized of step 2 and 3 achieves 89.7\% and 83.2\% separately for NMNIST (Table \ref{architecture_nmnist}) and FSDD (Table  \ref{architecture_fsdd}), which can reach or even surpass the pure three parallel structure (89.8\% for NMNIST, and 82.7\% for FSDD) \cite{wijesinghe2019analysis} without optimization of step 2 and 3, which indicates the effectiveness of our three-step search. This may because NMNIST is the image dataset of DVS output, which is more conducive to generating spike of distinguishing features for multi-layer structures of LSM, while FSDD is the speech dataset, and LSM is inherently easier to classify it. When the number of $N_{group}$ is more than five, our architecture space can take advantage of the benefits from parallel \cite{wijesinghe2019analysis} and series \cite{mi2019spatiotemporal} liquids. After optimization of neuron number and parameters for the two-layer eight liquids, it is easy to find a better model than pure three parallel best architecture on FSDD and NMNIST. 

Although the accuracy of eight liquids with a two-layer hierarchical architecture is no better than three parallel architecture when both of them are optimized for step 2 and 3, fewer connections are realized with more $N_{group}$ which can alleviate the memory and computation requirement. For example, the connections of a single LSM with $N_{total}$ neuron number can be expressed as $N^{2}_{total}$. 
Suppose a large liquid with the number of neurons $N_{total}$ is divided into $N_{group}$ smaller liquids, and the number of each small liquid is $N_{total} / N_{group}$. Then connections of multi-liquid is about $(N_{total} / N_{group})^{2} \times N_{group} = (N_{total})^{2} / N_{group}$. This shows that the number of connections will be reduced by $N_{group}$ times when the large liquid is divided into $N_{group}$ small liquids with the same percentage connectivity, which is friendly for memory-limited systems.
Compared to pure three parallel architectures, the connections within eight small optimized liquids with two-layer hierarchical architecture for NMNIST and FSDD is reduced by 56.2\% and 56.6\% respectively, which helps to reduce storage requirement for hardware.

In conclusion, our three-step search can maximize the advantages of LSM model in architecture, neuron number, and parameter space. In terms of accuracy for different datasets, the parallel architecture may be the best structure for the three datasets. Meanwhile neuron number variation (step 2) and parameter search (step 3) contribute to improve the accuracy of the specified architecture in different datasets. Besides, our three-step search is easier to find the optimal model with fewer connections (larger liquids number) and comparable accuracy for LSM on DVS and speech datasets such as NMNIST and FSDD.  



\paragraph{Neuron number minimizing}
\begin{figure}[H]
\centering
\includegraphics[width=0.9\linewidth]{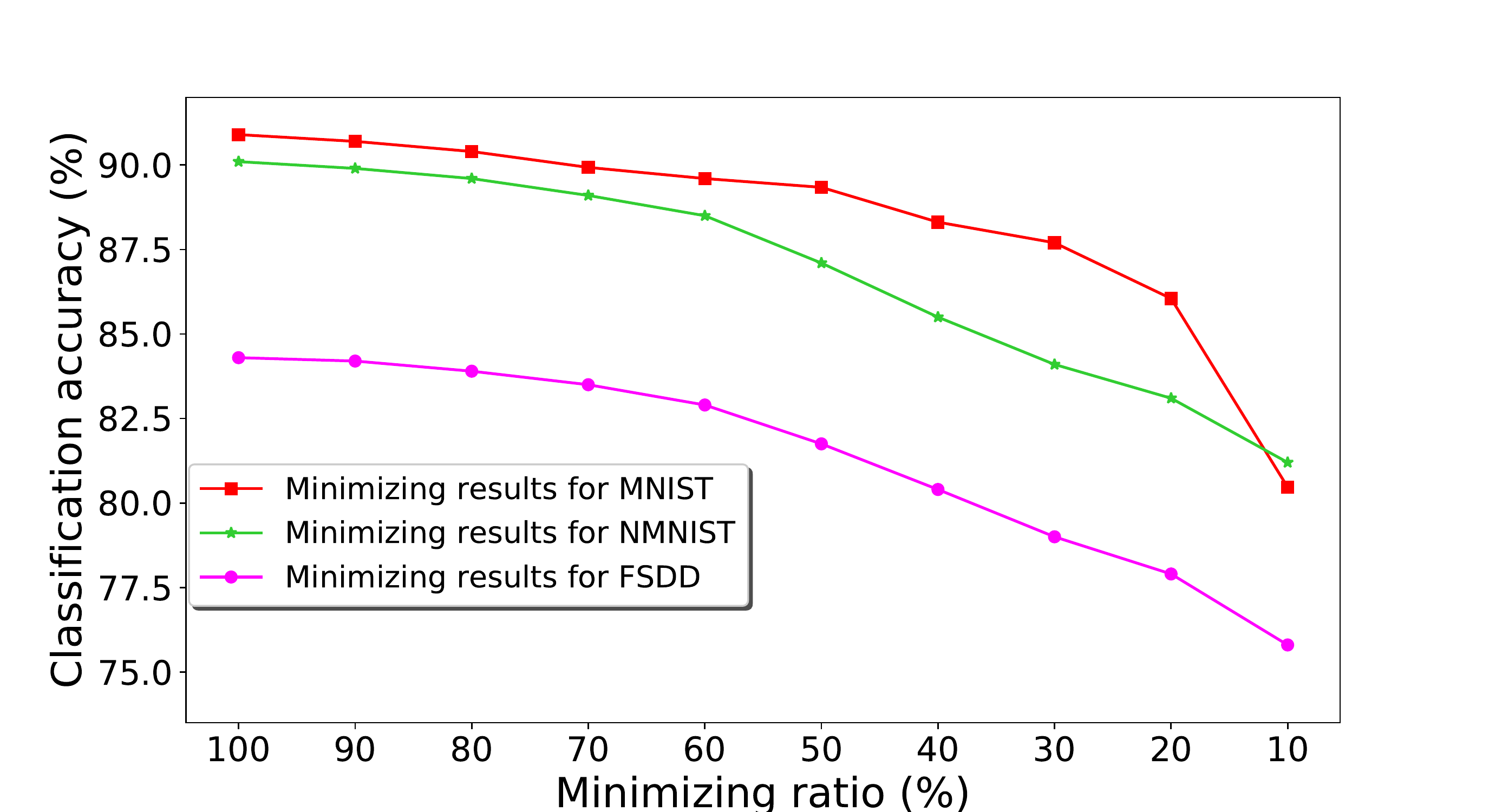}
\caption{The minimizing neuron number results on three datasets.}
\label{minimizing_results}
\end{figure}

To further reduce the hardware overhead, we perform a neuron number minimizing procedure to reduce the redundant neurons of the optimized LSM model (Table \ref{architecture_mnist}, \ref{architecture_nmnist1}, \ref{architecture_fsdd1}) on three different datasets.
In our experiments, the same minimizing ratio is applied to each small liquid for the optimal LSM.
As shown in Figure \ref{minimizing_results}, the neuron number could be reduced by 20\% with only about 0.5\% accuracy loss on three datasets.
Additionally, the accuracy decrease becomes unacceptable after reducing more than 40\% of the total neuron number. This may because LSM is a chaotic system with redundancy. Reducing the number of neurons may not have a great impact on its ability for classification.

\paragraph{Comparison}

\begin{table}[H]
\renewcommand{\arraystretch}{1.7}
\caption{The state-of-art classification accuracy on MNIST dataset using LSM model}
\label{performance_mnist}
\begin{tabular}{p{0.35\linewidth}p{0.4\linewidth}c}
\hline
Model          & Architecture & Accuracy \\ \hline
Single LSM (this work)     & 1000 $\times$ 1 + SGD  & 92.7\%   \\
Five parallel liquids \cite{wijesinghe2019analysis} & 200 $\times$ 5 +  SGD  & 95.5\%  \\
D-LSM \cite{wang2016dlsm}  & 37 $\times$ 3 + pooling, 460 + STDP  & 97.2\%  \\
this work      & 382, 99, 388, 131 +  SGD & 93.2\%  \\ \hline
\end{tabular}
\end{table}

\begin{table}[H]
\renewcommand{\arraystretch}{1.7}
\caption{The state-of-art classification accuracy on NMNIST dataset using LSM model}
\label{performance_nmnist}
\begin{tabular}{p{0.4\linewidth}p{0.35\linewidth}c}
\hline
Model          & Architecture  & Accuracy \\ \hline
Single LSM (this work)     & $1000 \times 1$ + SGD & 90.5\%  \\
Five parallel liquids (this work)& $200 \times 5$ + SGD  & 90.7\%  \\
lonic liquid \cite{Iranmehr2019bio}  & $25 \times 25$ + p-delta-rule   & 91.48\%  \\
this work           & 75, 505, 420 + SGD  & 92.5\% \\ \hline
\end{tabular}
\end{table}

\begin{table}[H]
\renewcommand{\arraystretch}{1.7}
\caption{The state-of-art classification accuracy on FSDD dataset using LSM model}
\label{performance_fsdd}
\begin{tabular}{p{0.4\linewidth}p{0.35\linewidth}c}
\hline
Model          & Architecture & Accuracy \\ \hline
Single LSM (this work)   & $1000 \times 1$ + SGD & 80.1\%   \\
Five parallel liquids (this work) & $200 \times 5$ + SGD & 82.3\%   \\
lonic liquid \cite{Iranmehr2019bio}  & $25 \times 25$ + p-delta-rule & 72.2\%   \\
this work           & 134, 382, 484  +  SGD  & 84\%   \\ \hline
\end{tabular}
\end{table}
For presenting more comparison, we list the performance and structure of state-of-art LSM models on MNIST, NMNIST, and FSDD datasets in Table \ref{performance_mnist}, \ref{performance_nmnist}, and \ref{performance_fsdd}.
The accuracy of optimal LSM models found by our framework is better than the single liquid for all the three datasets. Meanwhile the network connections can be reduced by 67.7\%, 56.3\% and 60.2\% respectively on MNIST, NMNIST and FSDD with an average connection reduction of 61.4\%. 
On MNIST dataset, we find a model structure similar to the best architecture in Parami \textit{et al.} \cite{wijesinghe2019analysis}, which verifies the effectiveness of our NAS-based framework. The accuracy gap between our work and Parami \textit{et al.} \cite{wijesinghe2019analysis} may be due to the parameter optimization of SGD algorithm. D-LSM \cite{wang2016dlsm} achieves the highest accuracy, while complex architectures such as the pooling layer in deep learning will induce extra hardware overhead. 

As shown in Table \ref{performance_nmnist} and \ref{performance_fsdd}, we achieve the highest accuracy on the NMNIST and FSDD datasets. Although the neuron number of lonic liquid \cite{Iranmehr2019bio} is less than our optimal model, 
the neuron number of our optimal model can be further reduced with only a small loss of accuracy as shown in Figure \ref{minimizing_results}. 
The minimized model of 625 neurons in our optimal model can reach 90.9\% and 82.9\% on NMNIST and FSDD, respectively.


\section{Conclusion}
In this paper, we present a dataset-oriented NAS framework for designing high-performance Liquid State Machines.
 For the first time, the NAS method is employed to find the optimal model of LSM. In addition, we use a three-step design to explore the hierarchical LSM architecture with both parallel and cascade multi-liquids, neuron number and parameters of each liquid. 
Three datasets with different properties, including MNIST, NMNIST, and FSDD are used to test the effectiveness of our proposed framework.
Simulation results show that the three-step searched optimal model can achieve comparable accuracy on classification tasks of the three datasets.
Furthermore, our proposed framework can also be used to reduce the hardware overhead with an acceptable performance by neuron number minimizing. In the future, we will implement our proposed framework with more than 1000 neurons, where LSM models with higher performance are expected to be found.
\bibliographystyle{ieeetr}
 
\bibliography{Ref.bib}

\end{document}